\documentclass[letterpaper]{article} 
\usepackage{aaai24}  
\usepackage{times}  
\usepackage{helvet}  
\usepackage{courier}  
\usepackage[hyphens]{url}  
\usepackage{graphicx} 
\usepackage{natbib}  
\usepackage{caption} 
\usepackage{algorithm}
\usepackage{algorithmic}

\usepackage{booktabs}
\usepackage{multirow}

\usepackage{newfloat}
\usepackage{listings}
\usepackage{graphicx}
\usepackage{xcolor}

\DeclareCaptionStyle{ruled}{labelfont=normalfont,labelsep=colon,strut=off} 
\floatstyle{ruled}
\newfloat{listing}{tb}{lst}{}
\floatname{listing}{Listing}
\title{Towards LLM-guided Causal Explainability for Black-box Text Classifiers}
\author{
    Amrita Bhattacharjee, Raha Moraffah, Joshua Garland, Huan Liu
}
\affiliations{
    Arizona State University\\

    Tempe, Arizona USA\\
    \{abhatt43, rmoraffa, joshua.garland, huanliu\}@asu.edu
}

\usepackage{bibentry}

\begin{document}

\maketitle

\begin{abstract}

With the advent of larger and more complex deep learning models, such as in Natural Language Processing (NLP), model qualities like explainability and interpretability, albeit highly desirable, are becoming harder challenges to tackle and solve. For example, state-of-the-art models in text classification are black-box by design. Although standard explanation methods provide some degree of explainability, these are mostly correlation-based methods and do not provide much insight into the model. The alternative of \textit{causal explainability} is more desirable to achieve but extremely challenging in NLP due to a variety of reasons. Inspired by recent endeavors to utilize Large Language Models (LLMs) as experts, in this work, we aim to leverage the instruction-following and textual understanding capabilities of recent state-of-the-art LLMs to facilitate causal explainability via counterfactual explanation generation for black-box text classifiers. To do this, we propose a three-step pipeline via which, we use an off-the-shelf LLM to: (1) identify the latent or unobserved features in the input text, (2) identify the input features associated with the latent features, and finally (3) use the identified input features to generate a counterfactual explanation. We experiment with our pipeline on multiple NLP text classification datasets, with several recent LLMs, and present interesting and promising findings.

\end{abstract}

\section{Introduction}

\begin{figure}
    \centering
    \includegraphics[width=\columnwidth]{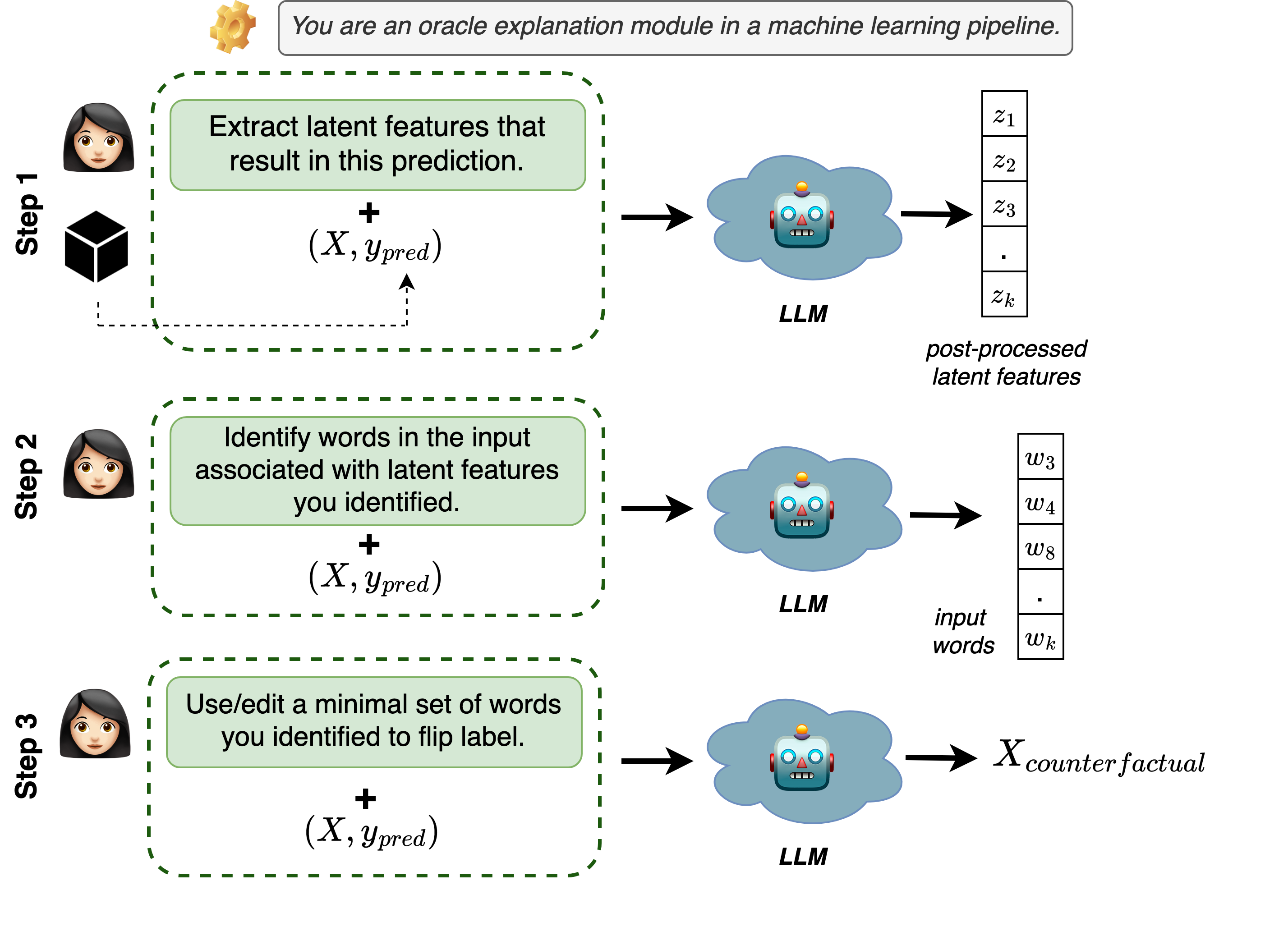}
    \caption{Our proposed three-step pipeline that leverages LLMs to (1) identify latent unobserved features, (2) identify input features associated with those latent features, and finally (3) use the input features to generate a counterfactual explanation.}
    \label{fig:framework}
\end{figure}

Deep learning models for NLP tasks, especially models built using transformers ~\cite{vaswani2017attention}, have achieved impressive performance on several well-established evaluation benchmarks~\cite{wang2018glue,wang2019superglue} and are therefore widely used in a variety of applications and downstream tasks. As the complexity of NLP tasks and evaluation benchmarks have increased over the years~\cite{zhou2020progress}, models with more complicated architectures and larger number of parameters have taken the spotlight, over `simpler' methods such as Bag-of-Words~\cite{qader2019overview} or TF-IDF~\cite{yun2005improved}. However, unlike most previous methods, these newer state-of-the-art models, with hundreds of thousands or even millions of parameters, are not interpretable by design. Furthermore, explainability of these black-box NLP models is crucial in industry use-cases in order to operationalize these models beyond standard evaluation benchmarks. However, explainability is still a fairly unsolved problem in many NLP tasks, such as in text classification~\cite{ribeiro2016should,camburu2018snli,liu2021explainaboard,balkir2022challenges}. The discrete nature of text makes it challenging to apply standard explainability methods that are otherwise somewhat well-established in domains like computer vision. Out of the various explainability methods explored by the community, methods such as attention scores~\cite{wiegreffe2019attention} are typically local and often do not generalize well~\cite{jain2019attention,balkir2022challenges}, and are also not applicable to black-box models where we only have query-access. Overall, most explainability methods in NLP build on top of correlations rather than causation: these models do not truly explain what features in the input `caused' the prediction. Different from correlation-based explainability methods, ones inspired by the causality literature~\cite{pearl2009causality} often explain models by comparing predictions of an input sample with that of its counterfactual. Therefore, in this work we focus on the task of text classification and aim to provide \textit{causal explainability} for black-box text classifiers.

However, identifying true causality in text is extremely challenging, again, due to the high-dimensionality, discrete nature and complexity of natural language~\cite{feder2022causal}. Some work in this direction have proposed to use counterfactuals for evaluating, and possibly explaining NLP models~\cite{wu2021polyjuice}. In the context of text classification, such \textit{counterfactual explanations} are essentially versions of the input text,  minimally modified so that the label predicted by the classifier changes. Counterfactuals are inherently generated based on a causal question: ``what could have been changed in the input so the decision of the model is flipped?". Given this underexplored area of counterfactual explanations for NLP, in this work we aim to investigate whether recent state-of-the-art instruction-tuned large language models (LLMs) can be used to generate counterfactual explanations to explain black-box text classifiers. 

To test this capability, we develop a pipeline consisting of a two-stage feature extraction step, followed by the counterfactual generation step. In the feature extraction step, we leverage the instruction-understanding and contextual understanding capabilities of the LLM to: (1) extract the latent unobserved features that `caused' the label, and then to (2) identify the set of input features, i.e., words, that are associated with the extracted latent features. The same language model is then leveraged to generate a counterfactual explanation by only modifying the words chosen via the two-step feature extraction, thereby generating high quality counterfactual explanations that try to preserve high semantic similarity with the original input. Unlike previous counterfactual generation work that performs local, word-level changes~\cite{wu2021polyjuice,madaan2021generate}, our method aims to identify the latent, unobserved features in the input text to generate causal explanations in the form of counterfactuals. In this exploratory study, we aim to investigate the following intriguing research question:


\textbf{RQ:} \textit{Can state-of-the-art large language models be leveraged to provide causal explainability in the form of counterfactual explanations, in the context of black-box text classification models?}



\section{Background \& Related Work}
\label{sec:related}


\paragraph{Causal and Counterfactual Explanations:}

Causal models of explainability offer theoretically grounded transparency and interpretability, and have also been used in many fairness related applications~\cite{beckers2022causal}. Such models and similar models based on theories of causality have been used in recommender systems ~\cite{xu2020learning, xu2021learning}, image classification ~\cite{o2020generative}, tasks involving tabular data ~\cite{white2019towards}, etc. Counterfactual explanations~\cite{molnar2020interpretable} are a category of causal explanations that are based on the question of what features in the input should be changed so that the decision of the model flips. This can translate to identifying a set of causal features that has causal relationships with the model's decisions~\cite{moraffah2020causal}. To this end ~\cite{karimi2021algorithmic, mahajan2019preserving} propose a series of causal counterfactual explanations for tabular data. Similar efforts have also been made in the context of behavioral and textual data ~\cite{ramon2020comparison}, as well as financial text data ~\cite{yang2020generating}. Despite some existing efforts in automatically generating counterfactual examples for NLP tasks ~\cite{madaan2021generate,wu2021polyjuice,robeer2021generating}, there is a noticeable gap in research on counterfactual explanations for explaining black-box text classifiers.

\paragraph{LLMs as Experts or Feature Extractors:} Large Language Models (LLMs), although initially intended for generating human-quality text, have become tremendously powerful and are being used in many other applications as well. Most recent LLMs are transformer-based models with billions of parameters~\cite{touvron2023llama,touvron2023llama2}, trained on enormous corpora of data~\cite{penedo2023refinedweb,gao2020pile}, and then further fine-tuned to enable instruction-following capabilities~\cite{ouyang2022training}. Some LLMs are further `aligned' with human preferences by training them with reinforcement learning from human feedback (RLHF)~\cite{christiano2017deep}. The size, scale and the intensive training on various types of data has enabled LLMs to be used in use-cases beyond mere text generation~\cite{bubeck2023sparks}. Recent work has explored the use of LLMs as data annotators~\cite{he2023annollm,bansal2023large}, information extractors ~\cite{wei2023zero,shi2023chatgraph}, and even as experts ~\cite{xu2023expertprompting}, etc. Inspired by these efforts, in this work, we explore and investigate the possibility of leveraging LLMs for causal explainability in a structured manner.

\section{Method: Extracting Counterfactual Explanations via LLMs}
\label{sec:flare}

We aim to explain decisions from a black-box text classifier by leveraging the power of LLMs. In this section, we describe our proposed method to generate counterfactual explanations by prompting off-the-shelf instruction-tuned LLMs. Our overall pipeline is shown in Figure \ref{fig:framework}. We follow a 3-step approach to perform the counterfactual explanation generation: 

\textbf{Step 1}: Given the input text and the predicted label from a black-box classifier, we prompt a LLM to extract the latent or unobserved features that led to the prediction.

\textbf{Step 2:} Next, we use the same LLM to identify the words in the input text that are associated with the set of latent features that it identified in the previous step.

\textbf{Step 3:} Finally, we leverage the generation capabilities of the LLM to generate the counterfactual explanation by editing the input text by changing a minimal set of the identified words.

In the following part, we describe these steps and components in more detail.

\subsubsection*{Black-box Classifier}

For the black-box text classifier we are aiming to explain, we use a pre-trained DistilBERT~\cite{sanh2019distilbert} model that is further fine-tuned on the specific task dataset. For the test set of each task dataset $(X_{test}, Y_{test})$, we simply extract the predicted label $y_i$ for each text $x_i \in X_{test}$ in the test set and form tuples of the form (input text, predicted\_label). We retain correctly classified samples, i.e. samples where $y_i = \hat{y_i}$, where $\hat{y_i}$ is the ground truth for input $x_i$, since these are the predictions we would want to explain using our method. This black-box classifier is used during evaluation of the generated counterfactual explanations again, since we want to test the effectiveness of the generated counterfactuals. The modular nature of our methods allows the use of any black-box classifier instead of DistilBERT.


\subsubsection*{Feature Identification using LLMs}

In this step, we aim to identify the features in $x_i$ that resulted in the correctly predicted label $y_i$. For the correctly classified test samples, we design prompts to extract features from the input text $x_i$ in two phases: first by extracting the latent features, i.e., higher level features such as `inconsistency', `ambiguity', etc. Then, in the second stage, we prompt the LLM to extract the input features i.e., words in the input that are related to the high level latent features extracted in the previous step. 

\subsubsection*{Input Modification to Generate Counterfactual Explanations}

Finally, we use the LLM to change a minimal set of words in the input to produce a counterfactual explanation, i.e., an edited version of the input that would change the label when fed into the same black-box classifier.

This two-step feature extraction followed by generation step should effectively leverage the latent or unobserved features in the text to generate causal explanations in the form of counterfactuals. This is different from prior counterfactual generation work that simply performs local, word-level changes. The prompts we use to enable this three-step pipeline are as follows:

\vspace{1.5mm}
\noindent\fbox{%
    \parbox{0.95\columnwidth}{%
\textbf{Step 1:} ``You are an oracle explanation module in a machine learning pipeline. In the task of \textcolor{blue}{[task description]}, a trained black-box classifier correctly predicted the label \textcolor{blue}{[$y_i$]} for the following text. Think about why the model predicted the \textcolor{blue}{[$y_i$]} label and identify the latent features that caused the label. List ONLY the latent features as a comma separated list, without any explanation. Examples of latent features are `tone', `ambiguity in text', etc.\\
---\\
Text: \textcolor{blue}{[input\_text]} \\
---\\
Begin!"
\\
\hrule
\vspace{1.5mm}
\textbf{Step 2:} ``Identify the words in the text that are associated with the latent features: \textcolor{blue}{[latent features]} and output the identified words as a comma separated list."\\\hrule
\vspace{1.5mm}
\textbf{Step 3:} ``\textcolor{blue}{[list of identified words]} \vspace{1mm}\\  Generate a minimally edited version of the original text by ONLY changing a minimal set of the words you identified, in order to change the label. It is okay if the semantic meaning of the original text is altered. Make sure the generated text makes sense and is plausible. Enclose the generated text within \textless new\textgreater tags."}}
\vspace{1.5mm}

The texts in \textcolor{blue}{blue} are variable for each input text.

\begin{table*}[]
\centering
\resizebox{\textwidth}{!}{%
\begin{tabular}{@{}cccccccccc@{}}
\toprule
\multirow{2}{*}{\textbf{LLM}} & \multicolumn{3}{c}{\textbf{IMDB}} & \multicolumn{3}{c}{\textbf{AG News}} & \multicolumn{3}{c}{\textbf{SNLI}} \\ \cmidrule(l){2-10} 
 &
  \textbf{LFS} $\uparrow$ &
  \textbf{Sem. Sim. }$\uparrow$ &
  \begin{tabular}[c]{@{}c@{}}\textbf{Token-level} \\ \textbf{dist.} $\downarrow$\end{tabular} &
  \textbf{LFS} $\uparrow$&
  \textbf{Sem. Sim.} $\uparrow$&
  \begin{tabular}[c]{@{}c@{}}\textbf{Token-level} \\ \textbf{dist.} $\downarrow$\end{tabular} &
  \textbf{LFS} $\uparrow$&
  \textbf{Sem. Sim.} $\uparrow$&
  \begin{tabular}[c]{@{}c@{}}\textbf{Token-level} \\ \textbf{dist.} $\downarrow$\end{tabular} \\ \midrule
text-davinci-003     & 97.2   & \underline{0.851}   & 0.237 & 37.2    & \underline{0.882}   & 0.166   & 39.2   & 0.839  & 0.202  \\
GPT-3.5              & 86.42  & 0.823  & 0.440  & 36.71   & 0.875   & 0.132  & 34.14  & \underline{0.865}  & 0.170   \\
GPT-4                & \textbf{98.0}     & 0.816   & 0.415 & \textbf{84.8}    & 0.628   & 0.256   & \textbf{70.4}   & 0.854  & 0.175  \\ \bottomrule
\end{tabular}%
}
\caption{Performance of our pipeline with the three different LLMs. Best LFS values are highlighted in \textbf{bold} and best semantic similarity values are \underline{underlined}. Ideally we would want high values of both LFS and semantic similarity, although there is a trade-off.}
\label{tab:main-results}
\end{table*}

\section{Experimental Settings}
\label{sec:exps}

In this section, we briefly describe the datasets and LLMs used for exploring and evaluating our pipeline.

\subsubsection{Datasets} Since we focus our method on the task of text classification, we use three text classification datasets: (1) IMDB ~\cite{maas-EtAl:2011:ACL-HLT2011} for sentiment classication of IMDB movie reviews (Internet Movie Database). The labels are `positive' and `negative'; (2) AG News\footnote{https://huggingface.co/datasets/ag\_news} for news topic classification on short news articles. The labels here are `the world', `sports', `business' and `science/technology'; and finally (3) SNLI\footnote{https://huggingface.co/datasets/snli} (Stanford Natural Language Inference) for natural language inference task. The labels here are `entailment', `contradiction' and `neutral'. For this exploratory study, we use evaluate our pipeline on 250 samples from each dataset.

\subsubsection{LLMs} We use three LLMs from OpenAI: (1) \texttt{text-davinci-003}: which is an instruction-tuned version of GPT-3~\cite{brown2020language}; (2) GPT-3.5\footnote{https://platform.openai.com/docs/models/gpt-3-5}: which is the standard ChatGPT model as made available through the OpenAI API; and (3) GPT-4~\cite{openai2023gpt}. All three models are available through the OpenAI API. For all three LLMs, we use top\_p sampling with $p=1$, temperature $t=0.2$ and a repetition penalty of $1.1$.


\subsection{Evaluation Metrics}

Counterfactual explanations generated by the pipeline should be (1) effective, i.e., it should successfully flip the label of the classifier, and (2) content-preserving, i.e., should be as similar to the input as possible~\cite{madaan2021generate,wu2021polyjuice}. Following prior work~\cite{madaan2021generate}, in order to evaluate (1), we use the Label Flip Score (LFS), which is simply the percentage of generated counterfactuals that successfully flip the decision of the model and hence are successful. For evaluating (2), we use two metrics: one for similarity in the latent embedding space, that is computed as the inner product of the USE (Universal Sentence Encoder)~\cite{cer2018universal} embeddings of the input and corresponding counterfactual; and one for distance in the token space that is computed as the average Levenshtein distance~\cite{levenshtein1966binary} between the two strings.

\section{Results}
\label{sec:results}

In this section, we go over results of using our pipeline for identifying latent and input features and generating the counterfactual explanations.

\subsubsection{Effectiveness of Generated Counterfactual Explanations}

We report the main quantitative results of our framework in Table \ref{tab:main-results}. As mentioned previously, we report the Label Flip Score, semantic similarity and token-space distance. We see that among the three LLMs we evaluate for this task, GPT-4 performs the best in all three datasets. We do see a trade-off between the LFS values and content-preservation metrics, as has been reported in previous work as well~\cite{madaan2021generate}. We see that for the SNLI dataset (which involves more of a reasoning-type task), all the LLMs struggle to some extent. This is possibly due to the fact that LLMs have been known to struggle with reasoning-type tasks. Also, for the AG News dataset, we see that \texttt{text-davinci-003} and GPT-3.5 have significant room for improvement. This might be due to the task being a multi-class classification with four labels, thus confusing the LLM in the generation step (i.e., when we prompt the LLM to make sure the label changes - it has too many options to choose from, and no definitive way to understand which pair of label flips is the easiest to achieve). However, the impressive performance of GPT-4 implies there might be ways to involve LLMs in pipelines to facilitate causal explainability of models.

\subsubsection{Ablation: Effectiveness of Three-Step Pipeline}

We show a small ablation study in Table \ref{tab:ablation}. We compare results of the full pipeline with variants where (1) the latent feature extraction step is removed (i.e., Step 1 in Figure \ref{fig:framework}), and (2) both latent and input feature extraction steps are removed. We see that the full pipeline performs the best, implying that the latent or unobserved features identified actually help in identifying appropriate input features, which in turn affect the quality of the counterfactual explanation generated.
\begin{table}[]
\centering
\begin{tabular}{@{}ccc@{}}
\toprule
\multirow{2}{*}{\begin{tabular}[c]{@{}c@{}}\textbf{Framework} \\ \textbf{Variant}\end{tabular}} & \multicolumn{2}{c}{\textbf{LLM}} \\ \cmidrule(l){2-3} 
                          & \textbf{text-davinci-003} & \textbf{GPT-3.5}        \\ \midrule
Full pipeline             & \textbf{97.2}    & \textbf{86.42} \\
\textit{without} Step 1            & 95.78            & 78.52          \\
\textit{without} Step 1 and Step 2 & 95.39            & 59.19          \\ \bottomrule
\end{tabular}
\caption{Ablation results on the IMDB dataset. Values in the table are LFS scores - higher values are better. Best LFS scores are in \textbf{bold}.}
\label{tab:ablation}
\end{table}

\begin{table}[t]
\centering
\resizebox{\columnwidth}{!}{%
\begin{tabular}{@{}cc@{}}
\toprule
\textbf{Input Text} &
  \textbf{Latent Features Identified} \\ \midrule
\begin{tabular}[c]{@{}c@{}}A woman with a green\\ headscarf, blue shirt and \\ a very big grin. \\ The woman has been shot. \\ \textit{(Label: contradiction)}\end{tabular} &
  \begin{tabular}[c]{@{}c@{}}Inconsistency in text, \\ contradiction in events,\\  negative sentiment, \\ severity of action, \\ narrative coherence\end{tabular} \\\midrule
\begin{tabular}[c]{@{}c@{}}An old man with a package \\ poses in front of an advertisement. \\ A man poses in front of an ad. \\ \textit{(Label: entailment)}\end{tabular} &
  \begin{tabular}[c]{@{}c@{}}Subject consistency, \\ action consistency,\\  object consistency, \\ location consistency\end{tabular} \\\midrule
\begin{tabular}[c]{@{}c@{}}A young family enjoys feeling\\  ocean waves lap at their feet. \\ A family is out at a restaurant.\\  \textit{(Label: contradiction)}\end{tabular} &
  \begin{tabular}[c]{@{}c@{}}Location discrepancy, \\ activity mismatch, \\ context inconsistency\end{tabular} \\\midrule
\begin{tabular}[c]{@{}c@{}}Two teenage girls conversing \\ next to lockers. People\\  talking next to lockers. \\ \textit{(Label: entailment)}\end{tabular} &
  \begin{tabular}[c]{@{}c@{}}Subject consistency, \\ action consistency, \\ object consistency\end{tabular} \\ \bottomrule
\end{tabular}%
}
\caption{Examples of original input text with label, and extracted latent/unobserved features from the SNLI dataset where our pipeline generates successful counterfactual explanations (i.e., classifier label flips). LLM used here is GPT-4. }
\label{tab:latent}
\end{table}

\subsubsection{Quality of Latent Features Extracted}

Since the term `latent feature', in the sense we are using it, may not be easily understandable to an LLM, we look at the quality of the latent features extracted to make sure these are useful. We show some examples from the SNLI dataset in Table \ref{tab:latent} with the associated GPT-4-extracted latent features. These latent features all seem high quality and informative towards why the classifier predicted that specific label. This further strengthens the idea that LLMs \textit{can} be used for identifying such unobserved latent features and facilitate causal explainability of models.

\section{Conclusion}
\label{sec:conclu}

In this paper, we developed a pipeline to investigate whether we can leverage the impressive language understanding capabilities of LLMs to facilitate causal explainability via counterfactuals for explaining black-box text classification models. Our quantitative and qualitative findings suggest that this idea is quite promising, provided high quality LLMs (such as GPT-4) are used. The pipeline proposed in this work can be utilized to effectively prompt off-the-shelf LLMs to first identify the latent features causing the prediction, then identify the input features associated with the latents, and then finally generating a modified version of the input by making minimal edits to the list of identified input features. This promising venture into causal explainability paves the way for further exploration of LLMs in other types of causal explainability and broadly causality in NLP, hinting towards possible uses in causal inference, causal discovery, etc. 

\section{Acknowledgments}

This work is supported by the DARPA SemaFor project (HR001120C0123), Army Research Office (W911NF2110030) and Army Research Lab (W911NF2020124). The views, opinions and/or findings expressed are those of the authors.

\bibliography{references}

\end{document}